\documentclass{article}
\usepackage{spconf,amsmath,graphicx,hyperref,amssymb,amsfonts,cite,float,caption,multirow,booktabs,adjustbox}
\title{PMMD: A POSE-GUIDED MULTI-VIEW MULTI-MODAL DIFFUSION FOR PERSON GENERATION}

\name{Ziyu Shang$^{\star}$ \qquad Haoran Liu $^{\dagger\star}$ \qquad Rongchao Zhang $^{\ddagger}$ \qquad Zhiqian Wei $^{\dagger}$ \qquad Tongtong Feng $^{\S}$}
\address{$^{\star}$Harbin Institute of Technology, Shenzhen, China \\
$^{\dagger}$City University of Hong Kong, Hong Kong, China  \\
$^{\ddagger}$Peking University, Beijing, China  \\
$^{\S}$Tsinghua University, Beijing, China}

\begin{document}
\ninept

\maketitle

\begin{abstract}
Generating consistent human images with controllable pose and appearance is essential for applications in virtual try on, image editing, and digital human creation. Current methods often suffer from occlusions, garment style drift, and pose misalignment. We propose Pose-guided Multi-view Multimodal Diffusion (PMMD), a diffusion framework that synthesizes photorealistic person images conditioned on multi-view references, pose maps, and text prompts. A multimodal encoder jointly models visual views, pose features, and semantic descriptions, which reduces cross modal discrepancy and improves identity fidelity. We further design a ResCVA module to enhance local detail while preserving global structure, and a cross modal fusion module that integrates image semantics with text throughout the denoising pipeline. Experiments on the DeepFashion MultiModal dataset show that PMMD outperforms representative baselines in consistency, detail preservation, and controllability. Project page and code are available at \url{https://github.com/ZANMANGLOOPYE/PMMD}.

\end{abstract}


\section{Introduction}
\label{sec:intro}
Pose guided person image generation aims to synthesize realistic images conditioned on a reference image and a target pose, while preserving the subject's identity and aligning with the target pose\cite{C1,C2,C3}. This technique is valuable for virtual try on, digital human animation, electronic commerce display, and person re identification. However, large discrepancies between the source image and the target pose make it challenging to produce plausible results when only these two conditions are available.

Early approaches were based on Generative Adversarial Networks\cite{C4,C5,C6} and Variational Autoencoders\cite{C7}. Although they achieved initial progress, these models often suffered from unstable training, limited fine detail, and inaccurate pose alignment. Recent research\cite{C21,C22} has shifted toward diffusion models, which synthesize targets through multi step denoising and thereby improve both stability and fidelity. For example, PIDM\cite{C1} introduces a texture diffusion module that embeds source features into a diffusion U-Net\cite{C19,C20} to enhance appearance fidelity. PoCoLD\cite{C8} leverages three dimensional DensePose annotations together with appearance pose interaction to strengthen consistency between target pose and identity. PCDMs\cite{C9} adopt a three stage diffusion pipeline that first predicts global features, then generates a low resolution image, and finally refines local textures. IMAGPose\cite{C10} incorporates feature level, image level, and cross view attention to enable multi view inputs and multi pose generation, improving alignment and detail while preserving identity. Despite these advances, current methods still struggle with missing information caused by pose induced occlusions, which often leads to deviations in clothing details from user expectations.

In parallel, text to image models enable conditioning through natural language and further improve generation quality within latent diffusion frameworks such as LDM\cite{C11}. Compared with single modality inputs, multi modality methods integrate text, reference images, and structural signals as complementary conditions to enhance controllability and generalization. IP-Adapter\cite{C12} injects semantic cues from images to reinforce identity preservation. ControlNet\cite{C13} exploits structural signals such as pose or edges through controllable branches. T2I-Adapter\cite{C14} introduces lightweight adapters for efficient fusion and transfer of textual and visual conditions. UPGPT\cite{C15} explicitly incorporates text prompts into a multi modal diffusion framework for pose transfer, highlighting the utility of language in this task. Nevertheless, the inherent randomness of text generation can still undermine identity preservation and fine grained garment details, indicating the need for stronger cross modal alignment and more precise control.

\begin{figure}[t] 
    \centering
    \includegraphics[width=1\linewidth, trim=0.2cm 0.5cm 0cm 0.5cm, clip]{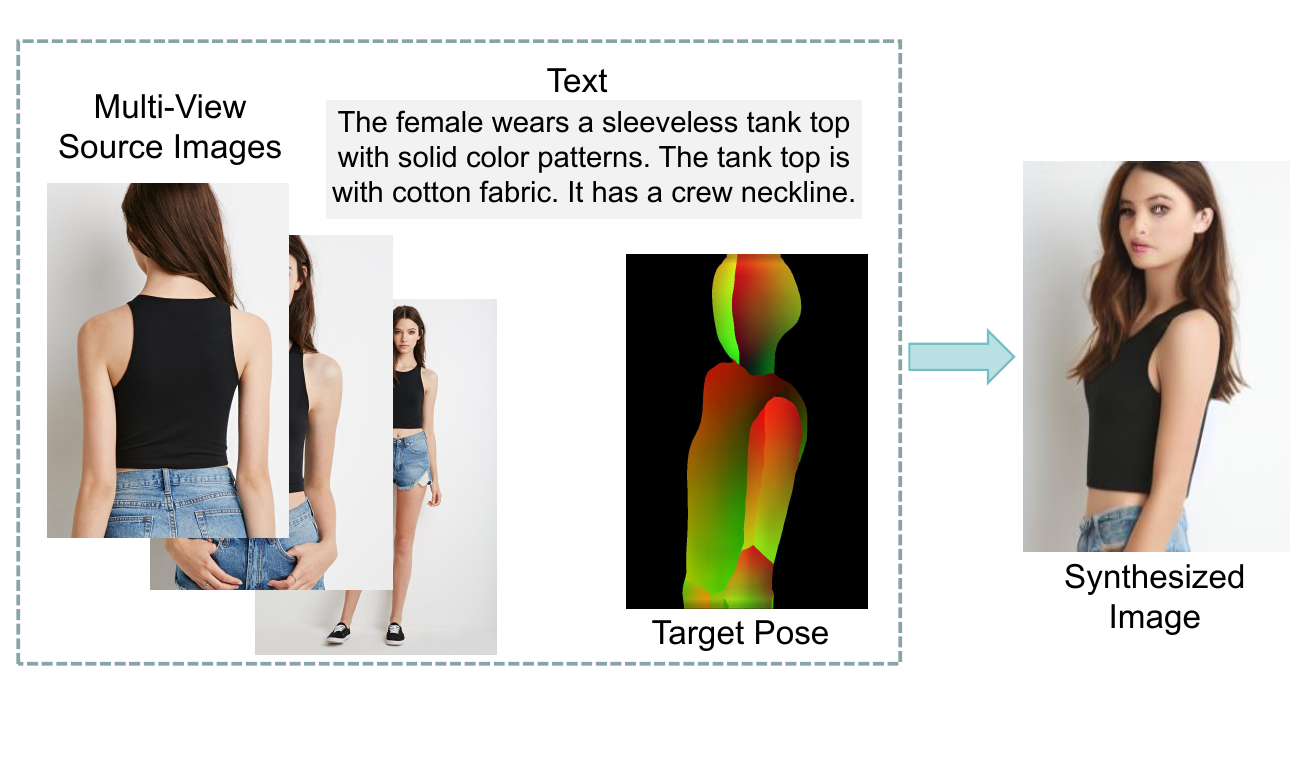}
    \caption{PMMD supports both text prompts and multi-view image inputs.}
    \label{fig:figure1}
\end{figure}
To address these challenges, we present PMMD (Pose-guided Multi-view Multimodal Diffusion), a diffusion framework that integrates a multimodal information encoder, a ResCVA module, and a cross modal fusion module for high fidelity and controllable multi view person image generation. The multimodal encoder jointly models multi view images, DensePose based pose maps, and text prompts, which reduces cross view inconsistency and stabilizes pose conditioned synthesis. In conditional modeling, a fine tuned cross modal fusion module aligns visual semantics with language through attention and conditions the denoising network to follow user instructions. To enhance detail preservation, ResCVA is inserted into the backbone to strengthen local textures while maintaining global structural coherence. Pose control is realized by a DensePose driven pose encoder, enabling precise and coherent body synthesis. To our knowledge, this is the first text driven multi view human image generation framework for pose transfer, offering a natural and efficient interface for controllable synthesis, as shown in Fig.~\ref{fig:figure1}. Experiments on DeepFashion-MultiModal\cite{C16} demonstrate state of the art performance. Quantitative and qualitative comparisons, together with user studies, show improved realism and superior cross view consistency.

Our main contributions are summarized as follows:
\begin{itemize}
    \item A text conditioned multi view diffusion framework for pose transfer that jointly exploits appearance, pose, and semantic cues from multi view images, DensePose maps, and text, effectively mitigating occlusion, limb distortion, and garment inconsistency.
    \item A ResCVA module for fine grained detail modeling that builds on residual cross view alignment, preserves global structure, and strengthens high frequency textures and edge details.
    \item A cross modal fusion module that aligns image semantics with language through attention and injects them into the diffusion process, improving semantic consistency and controllability of the generated results.
\end{itemize}

\section{PROPOSED METHODS}
\label{sec:methods}

\begin{figure*}[htbp]
    \centering
    \includegraphics[width=\textwidth, trim=1cm 1.5cm 1cm 1cm, clip]{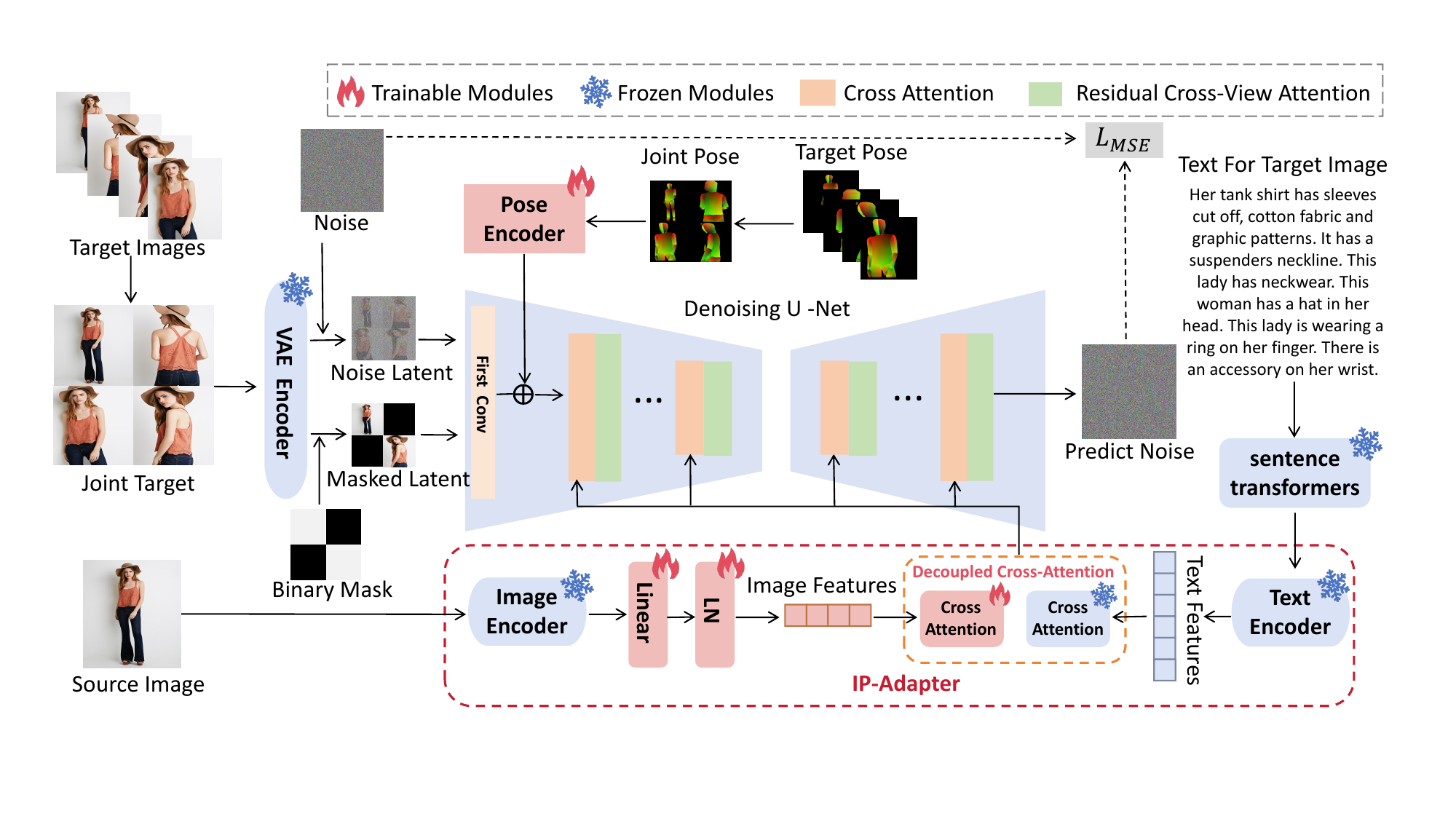}
    \caption{Overview of our model. PMMD consists of a multimodal feature encoder, a ResCVA module, and a cross-modal feature fusion module, enabling multi-view modeling. The multimodal feature encoder encodes the appearance image, pose map, and textual description, respectively. The fine-tuned cross-modal fusion module aligns image semantics with text prompts. Meanwhile, our proposed Residual Cross-View Attention (ResCVA) module enhances high-fidelity reconstruction of local details while preserving overall structural consistency.}
    \label{fig:figure2.pdf}
\end{figure*}

Our model is designed to achieve multi-view human image synthesis with textual prompts, enabling realistic generation by flexibly controlling the subject’s visual appearance, body pose, and textual description. The framework effectively addresses common challenges such as occlusions, severe distortions at limb extremities, and style inconsistencies in clothing, while meeting user demands for multi-view information and simple textual guidance. The overall architecture is illustrated in Fig.~\ref{fig:figure2.pdf}. The PMMD framework consists of a multi-modal feature encoder, a cross-modal feature fusion module, a ResCVA module, and a denoising backbone network. The input images from multiple views are combined into a joint image and encoded by a VAE encoder to obtain latent representations. These representations are further perturbed with Gaussian noise and combined with masked versions (covering unknown and target regions) to form noised and masked latent variables, thereby providing multi-view appearance information of the subject. Pose features are extracted from DensePose maps using a fine-tuned pose encoder to provide detailed pose control during generation. One source view is randomly selected as the image prompt and encoded by an image encoder. Textual prompts are first compressed using a lightweight Sentence-BERT encoder\cite{C18} and then passed through a text encoder. The fine-tuned cross-modal feature fusion module fuses textual and visual information and injects it into the cross-attention layers of the denoising U-Net\cite{C11}, enabling the model to better capture image-text correspondences. Finally, the ResCVA module is embedded to balance global context with local detail representations, leading to high-fidelity texture generation.
\vspace{0pt}
\subsection{Multimodal Feature Representation}
\label{ssec:mfr}
\textbf{Image Latent.}
We combine the multi-view source images and the target image into a $2 \times 2$ joint image \( x_D \in \mathbb{R}^{2H \times 2W \times C} \), which is then fed into the VAE encoder to obtain its latent representation:   \(E_I(x_D) = z \in \mathbb{R}^{\tfrac{2H}{f}\times\tfrac{2W}{f}\times d_V},\) where \( d_V \) denotes the channel dimension of the VAE and \( f \) is the downsampling factor (typically a power of $2$). Subsequently, Gaussian noise and a binary mask are applied. During training, $1–3$ views are randomly masked to simulate reference settings for single view and multiple views. At inference time, the trained diffusion model generates a new latent variable \(\hat{z}\), which is decoded by the VAE decoder back into the pixel space image:\(\hat{x}_D = D_I(\hat{z}).\)
A smaller downsampling factor \( f \) (e.g., \( f = 4 \)) preserves higher spatial resolution but significantly increases the latent space size and computational cost. Conversely, a larger \( f \) reduces computation but may cause detail loss during reconstruction, such as blurred facial regions in full-body images.\\
\textbf{Densepose.}
We adopt DensePose as the pose condition. Compared to sparse representations such as keypoints, DensePose provides pixel-level body part mappings, offering stronger structural constraints and finer-grained semantic representation. To effectively integrate this information into the diffusion model, we leverage ControlNet\cite{C13} to inject DensePose features into the generation process, thereby enabling precise pose control and structural consistency without compromising the capabilities of the pretrained model.\\
\textbf{Source Image.}
We leverage the CLIP\cite{C17} image encoder to extract clothing and appearance features. Compared with conventional convolution-based feature extractors, CLIP\cite{C17} is pretrained on large-scale image–text pairs and thus possesses stronger semantic alignment capabilities. This enables it to capture fine-grained clothing textures as well as holistic style information from the source image. During training, these features are injected into the diffusion model through the IP-Adapter module in the form of extended visual-semantic tokens, which helps preserve the clothing appearance and semantic consistency of the source image during pose transformation.\\
\textbf{Style Text.}
We adopt the clothing descriptions provided by the dataset as style text prompts. Since the original descriptions are relatively long and may exceed the input length limit of the CLIP\cite{C17} text encoder, we first employ a compact Sentence-BERT\cite{C18} variant to compress and distill the semantic content, producing summary-like keywords that preserve the essential information. These compressed text features are then encoded by the CLIP\cite{C17} text encoder into semantic vectors, which are fused with the source image features within the IP-Adapter\cite{C12} module. In this way, both textual semantic guidance and visual appearance information are jointly injected into the diffusion model, enabling more precise control over clothing style and semantic consistency.
\subsection{Residual Cross-View Attention Module}
\label{ssec:rescva}
To better capture both global and local dependencies in human image generation, IMAGPose\cite{C10} proposed the Cross-View Attention (CVA) module. This module is embedded after the cross-attention layers of the U-Net\cite{C11}, where the global feature is divided into multiple local features. By introducing an additional dimension, self-attention is learned among these local features, which are then reorganized into global features, thereby balancing global consistency and local detail modeling.

\begin{figure}[H] 
    \centering
    \includegraphics[width=1\linewidth, trim=3cm 4cm 7cm 7cm, clip]{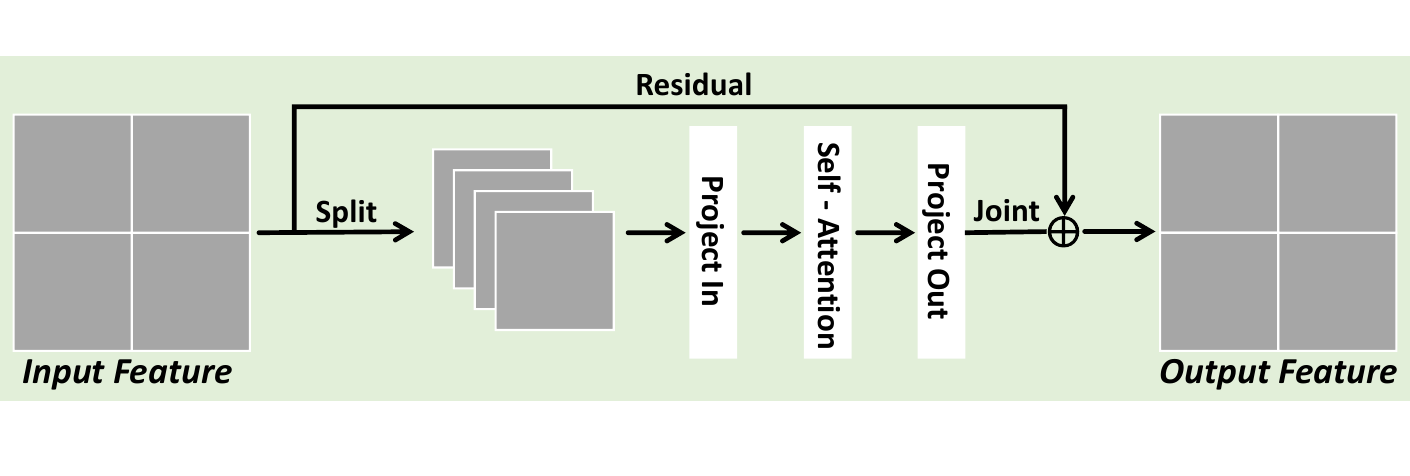}
    \caption{Illustration of the ResCVA module.}
    \label{fig:figure3.pdf}
\end{figure}

Building upon this, we further propose Residual Cross-View Attention (ResCVA, Fig.~\ref{fig:figure3.pdf}), which explicitly introduces a residual connection at the output of CVA:\begin{equation}
y = x + \mathrm{CVA}(x).
\end{equation}
where \(x\) denotes the input feature, \(\mathrm{CVA}(\cdot)\) represents the cross-view attention mapping function, and \(y\) is the output feature. The residual connection ensures that the cross-view enhanced features are incrementally superimposed on the original features.This design not only alleviates gradient vanishing during deep training but also enhances the preservation of original features, effectively improving training stability. As a result, the generated images achieve superior structural consistency and finer detail fidelity.
\vspace{-10pt}
\subsection{Training and Inference}
\label{ssec:train}
After multimodal feature encoding, we feed the representations into a modified denoising U-Net\cite{C11} of Stable Diffusion 1.5, while keeping the parameters of the pretrained diffusion model frozen. Specifically, the input layer of the U-Net\cite{C11} is extended to $8$ channels to jointly receive the concatenated noised latent and masked latent representations. Meanwhile, pose features are injected via ControlNet\cite{C13} to guide downsampling and intermediate residuals; text and image features are provided as conditional embeddings to the cross-attention layers, where our proposed ResCVA module is integrated to enhance both local and global feature interactions.

Accordingly, our loss function \(\mathcal{L}_{\text{MSE}}\) is defined as follows, where \(F_I\), \(F_T\), and \(F_P\) denote the image, text, and pose features, respectively:
\begin{equation}
\mathcal{L}_{\text{MSE}} = \mathbb{E}_{x_0, \epsilon, t, F_I, F_T, F_P} \left\| \epsilon - {\epsilon}_{\theta}(x_t, t, F_I, F_T, F_P) \right\|^2 
\end{equation}

We further apply random dropping of image conditions during training to enable classifier-free guidance at inference:
\begin{equation}
\begin{aligned}
&\hat{\epsilon}_{\theta}(x_t, F_I, F_T, F_P, t) = \\
&\quad \omega {\epsilon}_{\theta}(x_t, F_I, F_T, t)
+ (1 - \omega) {\epsilon}_{\theta}(x_t, F_P, t),
\end{aligned}
\end{equation}
Here, \(\omega\) denotes the guidance weight coefficient, which is used to balance the relative importance between the pose condition and the text-image condition.

\section{EXPERIMENTS}
\label{sec:experiments}
\textbf{Experimental Details.} We conduct experiments on the DeepFashion-MultiModal\cite{C16} dataset proposed by Text2Human, where each image is accompanied by a DensePose map and a textual description. The model is trained at a resolution of $256 \times 176$ using the Adam optimizer with a learning rate of $1$\(e\)-$5$ and a batch size of $1$. During training, image conditions are randomly dropped with a probability of $0.05$ to encourage the model to effectively learn the unconditional distribution. For sampling, we set the guidance weight \(\omega\) to $0.7$.\\
All experiments are carried out on a system equipped with eight NVIDIA TITAN RTX GPUs.
\subsection{Quantitative Comparison}
\label{ssec:quantitative}
We quantitatively compare our proposed PMMD with several state-of-the-art multimodal methods\cite{C12,C13,C14,C15}. For fairness, methods without pose guidance are integrated with PMMD by employing ControlNet\cite{C13} to encode DensePose. All methods are fine-tuned using the pretrained models provided by the original authors.

\begin{figure}[H]
    \centering
    \includegraphics[width=1\linewidth, trim=0cm 0cm 0cm 0cm, clip]{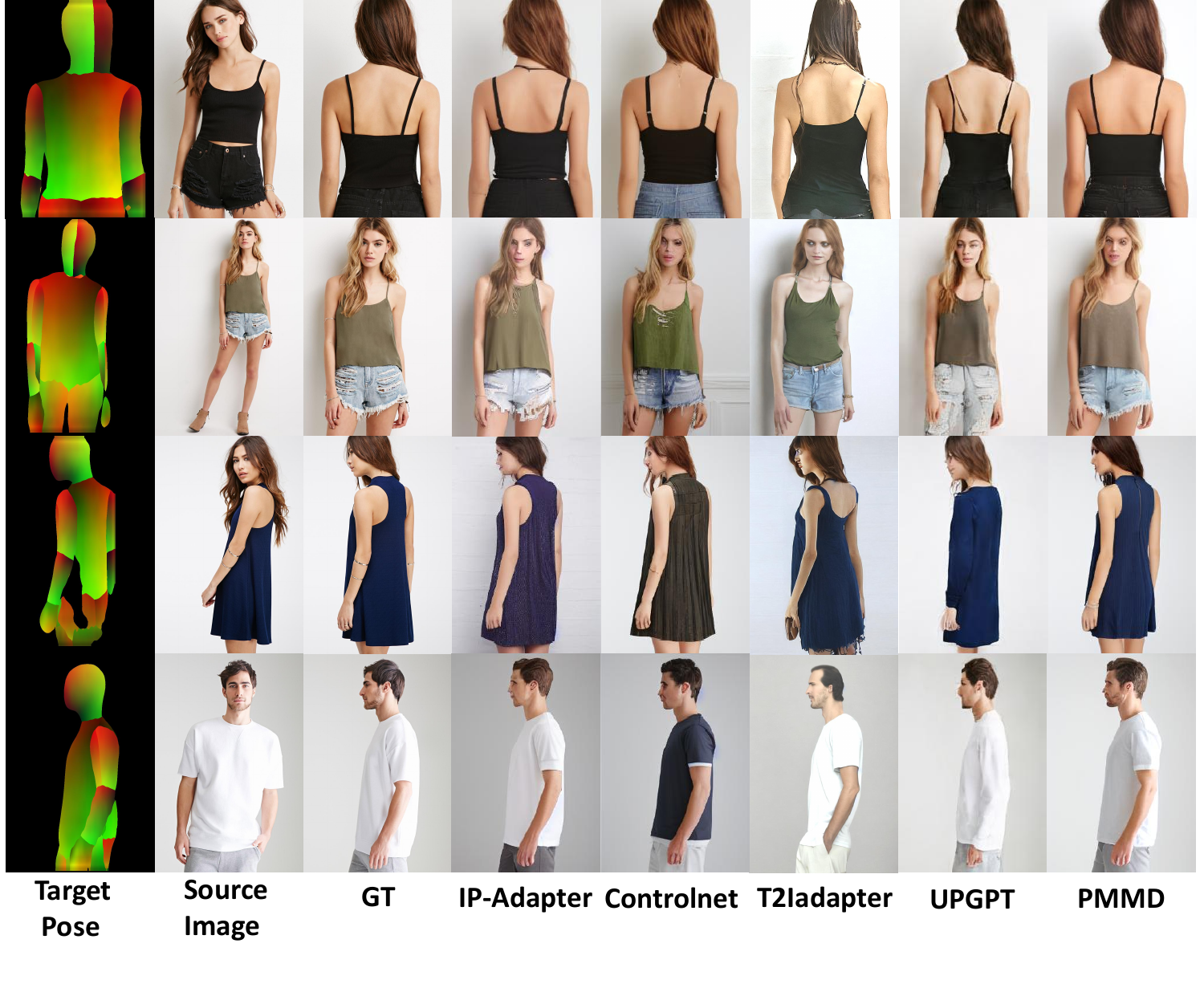}
    \caption{Qualitative comparisons with several state-of-the-art models on the
DeepFashion-MultiModal dataset.}
    \label{fig:figure_test.pdf}
\end{figure}

Table~\ref{tab:table1} presents the quantitative comparison on the DeepFashion-MultiModal\cite{C16} dataset. Our proposed PMMD achieves the best performance across all three metrics, with FID reduced to 8.5638, SSIM improved to 0.7397, and LPIPS decreased to 0.1909. Compared with existing methods, T2I-Adapter\cite{C14} and ControlNet\cite{C13} show limited improvements in SSIM and suffer from structural distortions; IP-Adapter\cite{C12} preserves appearance but remains suboptimal in texture details and perceptual consistency; and although UPGPT\cite{C15} attains a relatively high SSIM (0.7085), it still underperforms in perceptual quality and realism. In contrast, PMMD preserves pose alignment and produces clearer, more realistic textures, demonstrating the effectiveness of our design.
\vspace{-1em}
\begin{table}[H]
\centering
\caption{Quantitative comparison of the proposed PMMD with several state-of-the-art models in terms of SSIM, FID, and LPIPS.
Best results are in \textbf{bold}.}
\label{tab:table1}
\small
\adjustbox{max width=\linewidth}{%
\begin{tabular}{llccc}
\toprule
Dataset & Methods & SSIM $\uparrow$ & LPIPS $\downarrow$ & FID $\downarrow$ \\
\midrule
\multirow{5}{*}{\shortstack{DeepFashion- \\ MultiModal}}
 & T2Iadapter       & 0.4755 & 0.4107 & 20.573 \\
 & Controlnet       & 0.5263 & 0.3607 & 17.872   \\
 & IP-Adapter       & 0.6186 & 0.3004 & 13.032   \\
 & UPGPT            & 0.7085 & 0.2309 & 10.387  \\
 & \textbf{PMMD (Ours)} & \textbf{0.7397} & \textbf{0.1909} & \textbf{8.5638} \\
\bottomrule
\end{tabular}
}
\end{table}
\vspace{-1.0em}
Fig.~\ref{fig:figure_test.pdf} shows four representative results. In fact, IP-Adapter\cite{C12} has certain advantages in appearance preservation, managing to reproduce identity features to some extent, but the generated images suffer from blurry details, especially around edges and textures. ControlNet\cite{C13} achieves good pose alignment; however, due to insufficient appearance constraints, local features of the person often shift or become distorted. T2I-Adapter\cite{C14} performs reasonably well in pose generation, but its identity consistency is poor, as facial and clothing details frequently deviate from the source image. UPGPT\cite{C15}, while able to maintain some identity features, produces images of relatively low quality, with common issues such as edge blurring, disordered clothing structures, and missing textures, which reduce realism and stability. Notably, PMMD demonstrates superior performance, preserving fine texture details, maintaining accurate pose alignment, and reconstructing realistic body and facial features.
\subsection{User Study}
\label{ssec:usersdudy}
While quantitative evaluations offer objective evidence of the generated results, the assessment of human image synthesis is ultimately perception-driven. To this end, we conducted a user study with 90 participants to examine perceptual quality. The study had two parts: (a) comparison with real images, where 90 real and 90 generated samples were randomly selected, shuffled, and each shown for one second for authenticity judgment; (b) comparison with other methods, where 90 sets including a source image, target pose, ground truth, our result, and baseline outputs were presented, and participants selected the most realistic and convincing image. In this evaluation, R2G represents the proportion of real images judged as fake, G2R indicates generated images judged as real, and Jab measures the frequency of being rated as the best. Our approach achieved the best performance, demonstrating stronger alignment with human perception, as presented in Table~\ref{tab:userstudy}.
\vspace{-1em}
\begin{table}[H]
\centering
\caption{User study results comparing different methods. Best results are in \textbf{bold}.}
\label{tab:userstudy}
\small
\adjustbox{max width=\linewidth}{%
\begin{tabular}{lccc}
\toprule
Methods & R2G $\uparrow$ & G2R $\uparrow$ & Jab $\uparrow$ \\
\midrule
T2Iadapter & 22\% & 20.7\% & 6.7\% \\
Controlnet & 30.3\% & 32.3\% & 13.3\% \\
IP-Adapter & 34.5\% & 37.5\% & 16.7\% \\
UPGPT & 40.5\% & 47.0\% & 17.8\% \\
\textbf{PMMD (Ours)} & \textbf{44.3\%} & \textbf{55.4\%} & \textbf{45.6\%} \\
\bottomrule
\end{tabular}
}
\end{table}
\vspace{-1em} 
\subsection{Ablation Study}
\label{ssec:ablation}
We further evaluate our proposed method through ablation studies. Three variants are derived by selectively removing specific components from the full model (w/o ResCVA, w/o Text Summarization, and w/o mask). As shown in Table~\ref{tab:ablation}, the complete model achieves the best performance across all quantitative comparisons. These results highlight the contribution of each component, demonstrating that they are integral to the overall effectiveness of the method.
\vspace{-1em}
\begin{table}[H]
\centering
\caption{Ablation study on different components of our method. 
Best results are in \textbf{bold}.}
\label{tab:ablation}
\small
\adjustbox{max width=\linewidth}{%
\begin{tabular}{lccc}
\toprule
Methods & SSIM $\uparrow$ & LPIPS $\downarrow$ & FID $\downarrow$ \\
\midrule
base                   & 0.6186 & 0.3004 & 13.032  \\
w/o ResCVA             & 0.6737 & 0.2444 & 10.203  \\
w/o Text Summarization & 0.6955 & 0.2231 &  9.6746 \\
w/o mask               & 0.7271 & 0.2090 &  8.5638 \\
\textbf{PMMD (Ours)}   & \textbf{0.7397} & \textbf{0.1909} & \textbf{7.9580} \\
\bottomrule
\end{tabular}
}
\end{table}
\vspace{-1em}
\section{CONCLUSION}
\label{sec:typestyle}

In this work, we proposed PMMD, the first model that introduces textual prompts into the multi-view pose transfer task, establishing a unified multimodal diffusion framework. Through the synergistic integration of the multimodal feature encoder, the ResCVA module, and the cross-modal feature fusion module, the model achieves joint modeling of multi-view information, high-fidelity restoration of local details, and effective fusion of visual and textual semantics. In terms of pose control, the modeling of DensePose features with a pose encoder further enhances the refinement and controllability of human image generation. We demonstrate the effectiveness of PMMD for controllable human image synthesis through extensive quantitative evaluation, ablation studies, and user research. PMMD effectively alleviates the common issues in multi-view generation, including occlusion, garment style deviation, and pose instability, thereby providing a novel solution for human image synthesis under multimodal conditions.

\bibliographystyle{IEEEbib}
\bibliography{strings,refs}

\end{document}